\documentclass{article}

\usepackage{microtype}
\usepackage{graphicx}
\usepackage{subfigure}
\usepackage{booktabs} % for professional tables

\usepackage{hyperref}

\usepackage[accepted]{icml2019}

\usepackage{amsmath}
\usepackage{amsfonts}
\usepackage{mathtools}

\newcommand{\norm}[1]{\left\lVert#1\right\rVert}
\usepackage{bbm}

\icmltitlerunning{Numerically Recovering Critical Points}

\begin{document}

\twocolumn[
\icmltitle{Numerically Recovering the Critical Points of a Deep Linear Autoencoder}

\begin{icmlauthorlist}
\icmlauthor{Charles G. Frye}{rctn,hwni}
\icmlauthor{Neha S. Wadia}{rctn,biop}
\icmlauthor{Michael R. DeWeese}{rctn,hwni,biop,phys}
\icmlauthor{Kristofer E. Bouchard}{rctn,hwni,bse-lbnl,crd-lbnl}
\end{icmlauthorlist}

\icmlaffiliation{rctn}{Redwood Center for Theoretical Neuroscience, University of California, Berkeley, CA, USA}
\icmlaffiliation{hwni}{Helen Wills Neuroscience Institute, University of California, Berkeley, CA, USA}
\icmlaffiliation{biop}{Biophysics Graduate Group, University of California, Berkeley, CA, USA}
\icmlaffiliation{phys}{Department of Physics, University of California, Berkeley, CA, USA}
\icmlaffiliation{bse-lbnl}{Biological Systems and Engineering Division, Lawrence Berkeley National Lab, Berkeley, CA, USA}
\icmlaffiliation{crd-lbnl}{Computational Research Division, Lawrence Berkeley National Lab, Berkeley, CA, USA}

\icmlcorrespondingauthor{Charles Frye}{charlesfrye@berkeley.edu}

\vskip 0.3in
]

\printAffiliationsAndNotice{}

\begin{abstract}

Numerically locating the critical points of non-convex surfaces is a long-standing problem central to many fields. Recently, the loss surfaces of deep neural networks have been explored to gain insight into outstanding questions in optimization, generalization, and network architecture design. However, the degree to which recently-proposed methods for numerically recovering critical points actually do so has not been thoroughly evaluated. In this paper, we examine this issue in a case for which the ground truth is known: the deep linear autoencoder. We investigate two sub-problems associated with numerical critical point identification: first, because of large parameter counts, it is infeasible to find all of the critical points for contemporary neural networks, necessitating sampling approaches whose characteristics are poorly understood; second, the numerical tolerance for accurately identifying a critical point is unknown, and conservative tolerances are difficult to satisfy. We first identify connections between recently-proposed methods and well-understood methods in other fields, including chemical physics, economics, and algebraic geometry.
We find that several methods work well at recovering certain information about loss surfaces, but fail to take an unbiased sample of critical points.
Furthermore, numerical tolerance must be very strict to ensure that numerically-identified critical points have similar properties to true analytical critical points.
We also identify a recently-published Newton method for optimization that outperforms previous methods as a critical point-finding algorithm.
We expect our results will guide future attempts to numerically study critical points in large nonlinear neural networks.

\end{abstract}

\section{Introduction}

Neural networks have pushed forward the state of the art in a variety of machine learning tasks~\cite{schmidhuber2014,lecun2015}, but little is known about precisely how they work or why they can be effectively optimized to solutions with good test set performance.
In particular, it is not known why the non-convex training problem of a neural network is soluble using popular local methods such as stochastic gradient descent.
One approach to answering this question involves direct numerical interrogation of the loss surfaces of neural networks.
Efforts in this vein~\cite{pascanu2014,dauphin2014,choromanska2014,pennington2017} have demonstrated cases in which these loss surfaces enjoy a favorable ``no bad local minima'' property (that is, all minima are located at low values of the loss), and contain a proliferation of saddles instead, which was believed to be problematic for first-order methods.
Later theoretical work in non-convex optimization then showed that some stochastic first-order methods avoid saddle points of the training loss~\cite{lee2016,jin2018a} and converge to minima that generalize to the test loss~\cite{jin2018c}.
Taken together, these claims would do much to explain the trainability of neural networks.

However, a thorough understanding of the trainability of neural networks remains to be achieved. In particular, much is unknown regarding the properties of the loss surface, on which we focus in this paper. The theoretical results cited above concerning the shape of the loss surface are asymptotic and apply to simplified, approximate models~\cite{choromanska2014,pennington2017}.
The rates at which these asymptotic results apply and the degree to which these approximations break down as properties of the data and architecture are varied are unknown.
Furthermore, recent theoretical results~\cite{liang2018} have indicated that the loss surfaces for some common neural network architectures, e.g.~those with ReLUs and without skip connections, suffer from poor local minima even for linearly separable data.
Lastly, there is debate in the field on whether minima with different generalization performance exist and are characterized by their curvature properties~\cite{hochreiter1997,dinh2017,yao2018}.
Taken together, all these results underscore the need for further study of the shape of the loss surface.

One way to determine to what extent and why neural network loss surfaces can suffer from bad local minima, and what effect the distribution of minima, maxima, and saddle points (generically, points with zero gradient norm, or critical points) have on optimization and generalization, is to perform a thorough empirical examination of the interplay between architecture, data, and initialization strategy as they affect the loss surface.
In order to do so, robust numerical methods for finding critical points are needed.

Unfortunately, the problem of finding all of the critical points of a neural network loss surface is non-trivial. 
First, the rapid scaling of the number of critical points with input dimension~\cite{baldi1989,auer1996}, ignoring continuous equivalences~\cite{freeman2016}, is such that finding all of them can be impossible for any but the smallest of architectures.
This necessitates a sampling-based method, but the potential biases of such sampling methods are unknown and hard to quantify in the absence of ground truth.
Second, numerically finding points with gradient exactly equal to zero is unlikely with finite precision, necessitating the setting of a tolerance for the norm of the gradient.
This tolerance can be analytically determined for functions with Lipschitz conditions on the operator norm of the Hessian and on the norm of the gradient, but these values are poorly controlled for common neural networks~\cite{gouk2018} and worst-case upper bounds might be overly pessimistic in the typical case.
In the absence of ground truth, it is impossible to evaluate the efficacy of critical point-finding algorithms in solving these two problems.
Here we make progress in this direction by making a thorough study of three critical point-finding algorithms on a model where ground truth is available, thereby clarifying algorithmic choices that may yield accurate results in more complicated cases.

The problem of finding the critical points of neural network loss surfaces is actually a specific instance of an old, general problem in disguise: the problem of finding the zeros, or roots, of a function.
Here, that function is the gradient field of the loss surface.
In multiple other areas, e.g.~chemical physics~\cite{wales2005,ballard2017}, numerical algebraic geometry~\cite{sommese2005}, and economics~\cite{kehoe1987}, work extending back decades has identified algorithms for this task with varying convergence properties, domains of application, caveats, and scalabilities.

In Sec.~\ref{sec:single-cp}, we review methods used in previous papers that found the critical points of neural network loss surfaces in the context of this literature.
We also introduce a new algorithm, originally invented for another purpose, as a critical point-finding algorithm for neural networks.
We then apply these methods on a neural network loss surface where the ground truth identities of the critical points are known, that of a deep linear autoencoder~\cite{baldi1989}, which minimizes the same loss over the same class of functions as does Principal Components Analysis (PCA), but in a different parameterization.
Though linear networks do not have the same representational capacity as nonlinear networks, their training exhibits many of the properties of nonlinear network training~\cite{saxe2013}, and they provide a reasonably close test problem for algorithms of unknown efficacy.

We find that, while all the algorithms we study are capable of finding critical points (Fig.~\ref{fig:algorithm-performance}), strict cutoffs are necessary to ensure accuracy (Fig.~\ref{fig:cutoffs}), and all methods provide biased samples of the set of critical points (Figs.~\ref{fig:noise-sampling}~\&~\ref{fig:uniform-trajectory}).
We identify algorithmic choices that can improve performance and reduce this bias.

\section{Methods for Sampling Critical Points}
\label{sec:algorithms}

The problem of sampling the critical points of a loss surface requires two pieces to solve:
first, an algorithm capable of finding a single critical point;
second, a method for initializing this algorithm repeatedly in such a way that its outputs are a representative, ideally unbiased, sample of the true critical points.

\subsection{Preliminaries}\label{sec:definitions}
The loss surface $L$ is a scalar function of the $N$ parameters $\theta$ of the neural network,
We wish to understand the distribution of points $\theta^{*}$ where the gradient of the loss is zero by numerical means.
Formally, we define these, the critical points (CPs) of $L$, as the set 
$$\Theta=\{\theta^{*}:\nabla L\left(\theta^{*}\right)=0\}.$$
A CP $\theta^*$ can be classified by means of its (fractional) index $\alpha(\theta^{*})$, defined as the fractional number of negative eigenvalues $\lambda$ of the Hessian $\nabla^2L(\theta^{*})$ at that point: $\alpha(\theta^{*})=\text{index}(\theta^{*})=\sum_{\lambda}\mathbbm{1}(\lambda<0)/N$.
Note that when the Hessian is negative semidefinite $\alpha=1$ indicates a potential local maximum, $0<\alpha<1$ always indicates a saddle point, and when the Hessian is positive semidefinite $\alpha=0$ indicates a potential local minimum.

In practice, it is not possible to numerically locate points on $L$ where $\nabla L$ is identically zero.
Hence we introduce a relaxed definition of a CP, an $\varepsilon$-CP, defined as the set
$$\Theta_{\varepsilon}=\{\theta^{*}:\norm{\nabla L(\theta^{*})}^2<\varepsilon\}.$$
We will discuss in detail the effects of different choices of $\varepsilon$ on the results of running critical point-finding algorithms, shown in Fig.~\ref{fig:cutoffs}.

In the context of neural network loss surfaces, an often-studied property of $\Theta$ is the relationship of the indices of its members to their heights on the surface.
\cite{dauphin2014} and~\cite{pennington2017} proposed models of $\Theta$ and numerically found subsets of $\Theta_\varepsilon$ with the same loss-index relationship.
In this paper, we identify algorithmic choices that lead to recovery of a subset of $\Theta_{\varepsilon}$ with approximately the same loss-index relationship as $\Theta$ in a case where $\Theta$ is known.

\subsection{Finding a Single Critical Point}\label{sec:single-cp}

\subsubsection{Gradient Norm Minimization}

Given the problem of finding points that approximately satisfy a certain criterion, the natural optimization approach is to convert that criterion into a differentiable loss function and minimize it by local methods.
For the problem of finding points with small gradient norm, the result is an auxiliary loss function
\begin{equation}
    G(\theta) = \frac{1}{2}\norm{\nabla L(\theta)}^2.
    \label{eq:gnm-obj-fun}
\end{equation}
An instance of this class of methods was independently proposed for the problem of critical point-finding in neural networks in~\cite{pennington2017} for the first time, but in fact they have a long history in chemical physics, dating back to the 1970s under the name ``gradient norm minimization'' (GNM) ~\cite{mciver1972}, and being simultaneously and independently rediscovered thirty years later~\cite{angelani2000, broderix2000}.

There are two major concerns with this class of methods.
First, the problem is approximately quadratically worse-conditioned than the original problem~\cite{mciver1972}, and neural network losses are already poorly-conditioned~\cite{sagun2017}, resulting in very slow convergence for first-order methods.
Second, the loss surface of GNM, ironically, can suffer from a bad local minima property of its own, which is to say that it contains local minima that are not true critical points of the original loss surface.
These arise anywhere that the gradient lies in the nullspace of the Hessian.
It has been shown in the chemical physics literature that on some surfaces these spurious local minima dominate the global minima~\cite{doye2002}.

\subsubsection{Newton Methods}

Other approaches to finding points with zero gradient norm are better understood and have better convergence properties.
In particular, the zeros of the gradient field are solutions to a nonlinear system of equations, $\nabla L(\theta) = 0$, and can be found using root-finding techniques.
The classic root-finding algorithm is Newton-Raphson~\cite{izmailov2014}, which computes an update $\Delta\theta$ by solving the following linear system of equations:
\begin{equation}
    0 = \nabla^2 L(\theta) \Delta\theta + \nabla L(\theta).
    \label{eq:pure-newton}
\end{equation}
Though this algorithm enjoys rapid local convergence~\cite{boyd2004}, it has no global convergence guarantees on non-smooth functions, and the radius of local convergence can be zero if the Hessian is singular at the solution~\cite{griewank1983}.
Indeed, early work on finding critical points of neural network loss surfaces found that Newton-Raphson (with a fixed, non-unit step size) often failed to converge~\cite{coetzee1997}.
Newton-Raphson is therefore typically augmented with additional machinery to guarantee global convergence on a wider class of functions.
We consider two options in this paper.

The first, which we call Newton-TR, follows~\cite{dauphin2014} and uses a Levenberg~\cite{levenberg1944} scheme, equivalent to a trust region approach. Instead of solving Eq.~\ref{eq:pure-newton}, the  modified equation
\begin{equation}
    0 = \left(\nabla^2 L(\theta) + \gamma I\right) \Delta\theta + \nabla L(\theta),
    \label{eq:newton-tr}
\end{equation}
where $I$ denotes the identity matrix, is solved for multiple fixed values of $\gamma$.
In an optimization context, the update that results in the lowest value of $L$ is used.
In the context of root-finding, we instead take the update that results in the lowest value of $\norm{\nabla L}$.

The second is based on the recently proposed Newton-MR method~\cite{roosta2018}, for ``minimum residual''.
As above, Eq.~\ref{eq:pure-newton} is solved for $\Delta\theta$, which is then used as a line search direction with a novel set of conditions, derived by applying the Wolfe conditions to the squared gradient norm (see~\cite{roosta2018} for details).
This method was proposed as an optimizer for functions with the property that the gradient is only zero at minimizers and proven to converge to points with low gradient norm.
This makes it an appealing candidate for root-finding, unlike other Newton methods that are designed for more restricted classes of functions.

\subsubsection{Other Methods}

Here we review other methods for finding critical points of neural network loss surfaces and which provide promising targets for future research.

Eigenvector-following methods~\cite{trygubenko2004} are commonly used to find critical points of a desired index in chemical physics.
This is accomplished by initializing a quasi-Newton method, such as L-BFGS~\cite{liu1989}, at a local minimum and reversing the sign of the updates of that algorithm in a fixed set of directions at every step, corresponding to the desired index of the saddle being searched for.
These methods are primarily used to find low-index saddles, rather than all critical points, and require prior knowledge of a local minimum.
Eigenvector-following methods were applied to neural networks in~\cite{ballard2017} and~\cite{mehta2018a}.

Homotopy methods comprise another class of approaches for root-finding, most prominently in numerical algebraic geometry~\cite{sommese2005}, where the polynomial form of the nonlinearity can be exploited.
Homotopy methods use continuous transformations to deform solutions of a problem whose answers are known by construction into the solutions of a problem of interest~\cite{davidenko1953,broyden1969}.
They were first applied to neural network loss surfaces in~\cite{coetzee1997} and again more recently to linear network losses by~\cite{mehta2018b}.
The latter took advantage of the polynomial structure of squared losses applied to linear networks to use advanced methods from numerical algebraic geometry~\cite{bates2013}.
Outside of the case of polynomials, convergence guarantees are harder to come by.

\subsection{Sampling Multiple Critical Points}\label{sec:sampling-cps}

Given an algorithm that can find a single critical point, the next problem is to define a method for initializing this algorithm repeatedly in such a way that the outputs form a representative sample of the true critical points.
This presumes that the goal of examining the loss surface is to determine its mathematical properties, rather than the properties of, e.g., the parts of the loss surface which typical optimizers encounter from typical initializations.
Two heuristics have been used in previous work.
In both methods, the iterates of an optimization algorithm applied to the loss surface are used to propose points.
In the first, from~\cite{dauphin2014}, these points are sampled uniformly from the iterates.
We term this method ``uniform iteration''.
In the second, from~\cite{pennington2017}, iterates are sampled uniformly according to their height on the loss surface.
We term this method ``uniform height''.
As the identities of the critical points sampled by such a method are determined by the combination of loss surface shape, optimization algorithm behavior, and critical point-finding algorithm behavior, there is no guarantee of even sampling.
In an attempt to reduce any possible sampling bias,~\cite{dauphin2014} perturbed sampled points with additive noise.
We compare the sampling bias of both initialization methods in their noisy and noiseless versions below (see Figs.~\ref{fig:noise-sampling} \&~\ref{fig:uniform-trajectory}).

\begin{figure*}[ht!]
\centering
\includegraphics[width=0.9\linewidth]{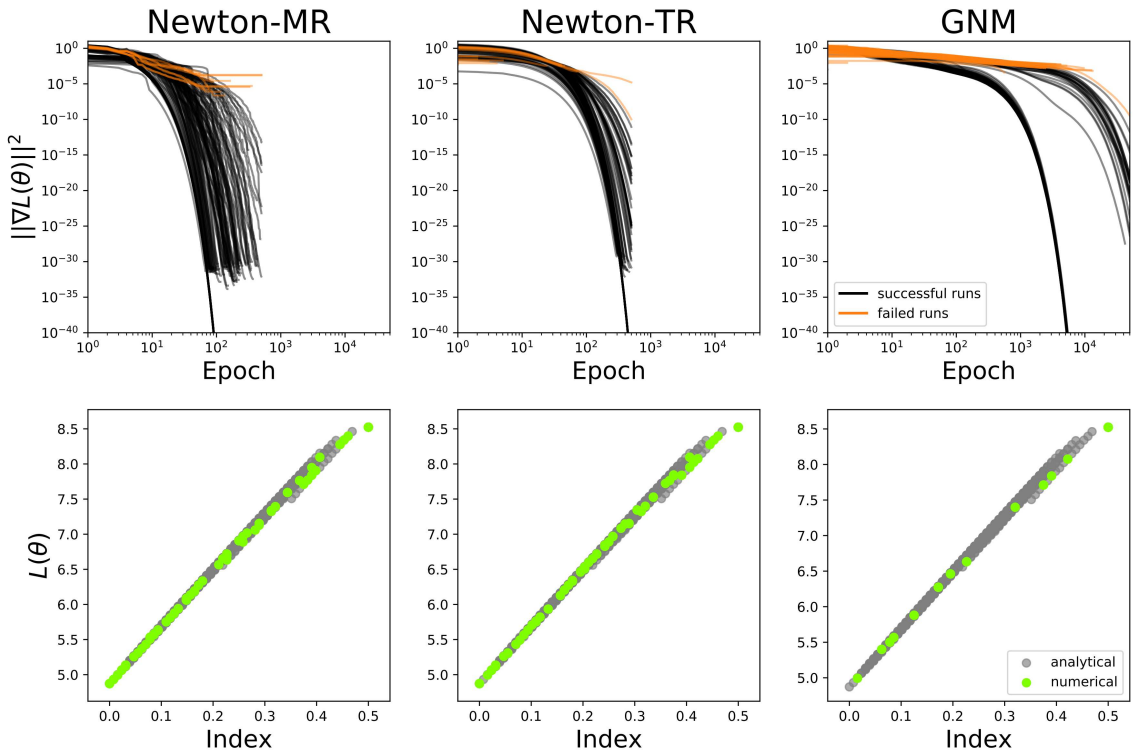}
\caption{\textbf{Newton-MR, Newton-TR, and GNM can recover critical points of a deep linear autoencoder}.
Top panels show squared gradient norms across epochs (both in log scale).
Black lines correspond to runs that terminated below a criterion value, while orange lines correspond to runs that terminated above it.
Due to the greater density of floating point numbers around zero, trajectories converging to the critical point at zero can achieve much lower squared gradient norms (as low as 1e-253); the y-axis is cut off at 1e-40 to focus on the other critical points.
Right panels show the loss and index for critical points found using numerical algorithms (in green) overlaid on the true values (in gray).
Each method was executed a total of 150 times: 10 optimization trajectories (each with a different random initialization) were used as seeds, with 15 initial points for each critical point-finding algorithm chosen at random from each trajectory.
}
\label{fig:algorithm-performance}
\end{figure*}

\section{The Deep Linear Autoencoder: A Model with Ground Truth}\label{sec:dlae}

Linear deep networks exhibit many of the training~\cite{saxe2013} and generalization~\cite{advani2017} complexities of their nonlinear counterparts while simultaneously being more amenable to analysis.
The loss surface of a feedforward single-hidden layer linear network with squared error loss was first studied in~\cite{baldi1989}, where the authors demonstrated that there are no non-global minima.
Recent work in linear networks has extended that result to multi-hidden layer networks~\cite{zhou2017} and to arbitrary differentiable convex losses~\cite{laurent2018}.
Unlike in nonlinear networks, the identity of every critical point is known in the single-hidden layer, linear case.
This provides a ground truth against which numerical results can be compared.
Here we study the performance of critical point finding algorithms on a single-hidden layer deep linear autoencoder (DLAE) with sixteen input and output units and four hidden units, applied to Gaussian data with a diagonal covariance matrix and evenly spaced eigenvalues between 1 and 16.
What follows is a description of the set of critical points $\Theta$ of the loss surface of this network, after~\cite{baldi1989}.

Each critical point of a DLAE with $n$ input units and $h$ hidden units ($h\leq n$) corresponds to a projection of the data onto a space spanned by some combination of at most $h$ of the $n$ eigenvectors of the data covariance matrix.
In the network parameterization, this $n\times n$ projection matrix (rank $\leq h$) is factored into the $h\times n$ and $n\times h$ input and output weight matrices.
This factorization is only unique up to an invertible linear transformation, and so each critical point is part of a disconnected Lie group of critical points.\footnote{This situation is also partially shared by ReLU networks~\cite{freeman2016}.} In the following, we only consider critical points modulo this equivalence relation, up to which the total number of critical points is given by the sum of the numbers of ways to choose from 0 to $h$ elements from an $n$ element set:
$\sum_{i=0}^h \binom{n}{i}$.
The loss surface of the $16\times 4\times 16$ DLAE we consider therefore contains a single minimum and $2516$ saddles.
The minimum corresponds to a solution where the network has learned to project onto the four eigenvectors with the largest eigenvalues, and the saddles correspond to all other projections.
Note that the single critical point that corresponds to learning none of the eigenvectors, resulting in a parameter vector of all zeros, shall be referred to as the ``critical point at zero'' later on.

It is possible to construct the weight matrices for the DLAE so as to initialize it to a particular critical point: the input-to-hidden weight matrix is constructed by placing the eigenvectors represented at that critical point in its columns, while the hidden-to-output weight matrix is its transpose.
This allows us to compute every element of $\Theta$, up to equivalence, and then compute the true values of, e.g.~the loss and index.
These values can then be compared to those computed from a subset of $\Theta_\varepsilon$, i.e.~critical points obtained via numerical methods.

\section{Results}\label{sec:results}

To find critical points, we first computed optimization trajectories by training a $16 \times 4 \times 16$ linear network on Gaussian data with the squared error loss.
These optimization trajectories were then used to generate initial points for critical point-finding algorithms.
Unless otherwise stated, initial points were selected uniformly with respect to height on the loss surface.
Critical point-finding algorithms were terminated either when no proposed step size met acceptance criteria or when a maximum number of epochs was reached.
Returned points were accepted as numerical critical points if their squared gradient norm was less than 1e-10.

To optimize the gradient norm objective (Eq.~\ref{eq:gnm-obj-fun}), we use batch gradient descent with back-tracking line search using the Wolfe conditions~\cite{wolfe1971}.
We use fast Hessian-vector products~\cite{pearlmutter1994} to compute our updates with the AutoGrad~\cite{maclaurin2016} Python package.

For both Newton methods, we use the robust linear solver \texttt{MR-QLP}~\cite{choi2011}.
See~\cite{roosta2018} for a succinct explanation of the benefits of this solver for poorly-conditioned, indefinite problems.
This solver is also accelerated by the use of fast Hessian-vector products.

See Supplementary Materials for further details and hyperparameter values.

\subsection{Newton Methods Outperform Gradient Norm Minimization}

$\varepsilon$-CPs found by all three numerical methods can have the same qualitative and quantitative loss and index values as do true CPs, but the methods have varying efficiencies (Fig.~\ref{fig:algorithm-performance}).
With the selected convergence criteria (see caption), none of the methods find $\varepsilon$-CPs in regions of the loss-index plane where there are no true CPs (e.g.~bad local minima, in the top left quadrant, or low-lying saddles, in the bottom right quadrant).
All methods also find subsets of $\Theta_\varepsilon$ that span the same values of loss and index as does $\Theta$.

Newton-MR (Fig.~\ref{fig:algorithm-performance}, left column) terminates in fewer iterations than does Newton-TR (Fig.~\ref{fig:algorithm-performance}, middle column; medians 221 and 430; Mann-Whitney $U$ = 4768.5, $p \ll 0.01$).
Newton-TR furthermore requires multiple calls to \texttt{MR-QLP} per iteration (one for each choice of trust region size), and so Newton-MR terminates more quickly (26.5 s per 100 iterations for Newton-MR vs 1 min 39 s for NewtonTR, on commodity hardware).
A more sophisticated mechanism for determining trust region size, rather than simply selecting the best choice from a pre-defined set, might narrow this performance gap.

Gradient norm minimization (GNM; Fig.~\ref{fig:algorithm-performance}, right column) is less efficient in two ways.
First, successful runs of GNM take on the order of one hundred times as many epochs to reach the same value of the gradient norm as do the Newton methods.
Each iteration only requires a single Hessian-vector multiply, unlike the Newton methods, whose calls to \texttt{MR-QLP} require several Hessian-vector multiplies, but the difference in wall time is still more than an order of magnitude (6.8 s per 100 iterations for GNM).
Furthermore, as discussed above in Section \ref{sec:algorithms}, GNM tends to get stuck in local minima of its objective function, as evidenced by the numerous short orange traces which terminate without the gradient norm going below $\varepsilon=$1e-10 (62.7\% of runs).
Because of this, even though the same number of runs and more compute were given to GNM, the number of $\varepsilon$-CPs recovered is far smaller (cf.~Fig.~\ref{fig:algorithm-performance}, bottom-right and bottom-left panels).
Note that local minima of the gradient norm objective do not correspond to local minima, or critical points at all, of the original problem.
Thus we conclude that Newton methods in general, and Newton-MR in particular, are a better choice of critical point-finding algorithm for neural network loss surfaces.

\subsection{Strict Cutoffs are Necessary to Accurately Recover Critical Points}

\begin{figure}[ht]
\centering
\includegraphics[width=\linewidth]{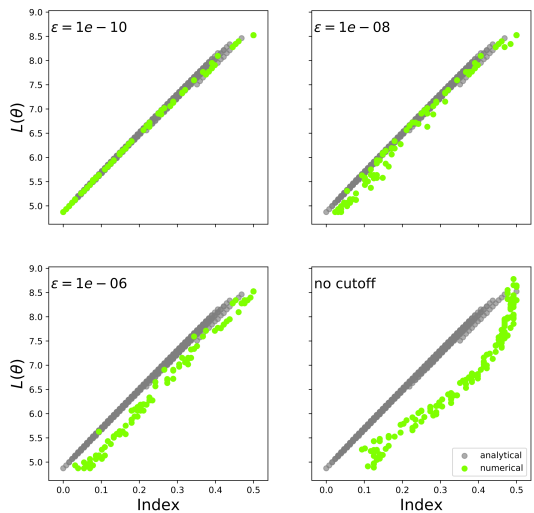}
\caption{\textbf{Cutoffs above 1e-10 are insufficient to guarantee accurate loss and index recovery}.
As in Fig.~\ref{fig:algorithm-performance},
$\varepsilon$-CPs are plotted in green over true CPs in gray.
For each panel, points are selected by taking the 150 runs of Newton-MR in Fig.~\ref{fig:algorithm-performance} and taking the first point whose squared gradient norm is below the cutoff value, $\varepsilon$, in the top-left corner.
For the bottom-right panel, we choose $\varepsilon=\inf$, which corresponds to accepting the initial point as an $\varepsilon$-CP.
}
\label{fig:cutoffs}
\end{figure}

Here and in previous work, the final output of a critical point-finding algorithm was accepted if the squared gradient norm on termination was below some cutoff value $\varepsilon$ (here, 1e-10).
However, at termination, many points may be far below this cutoff.
In fact, the vast majority of runs terminate with squared gradient norm close to 1e-30, and so the results in Fig.~\ref{fig:algorithm-performance} do not demonstrate that simply having squared gradient norm at the cutoff 1e-10 is sufficient to guarantee a match between the loss and index values of elements of $\Theta_\varepsilon$ and of $\Theta$.
If a sufficient cutoff could be identified, then much iteration time could be saved by terminating runs as soon as the squared gradient norm went below that value.

We found that, while having squared gradient norm equal to our cutoff was sufficient to guarantee accurate recovery of loss and index values, for the cutoff values used in~\cite{dauphin2014} and~\cite{pennington2017}, it was not (Fig.~\ref{fig:cutoffs}).
The error is larger for lower values of the loss.
Note, however, that the overall shape of the loss-index relationship is largely preserved for these looser cutoffs.
Interestingly, we find that simply computing the loss and index of points along the optimization trajectory results in the same concave-up shape reported in previous work~\cite{dauphin2014,pennington2017} (Fig.~\ref{fig:cutoffs}, bottom-right panel),
underscoring the need for care in the selection of convergence criteria.

\subsection{Adding Noise Reduces Sampling Bias}

\begin{figure*}[ht!]
\centering
\includegraphics[width=0.9\linewidth]{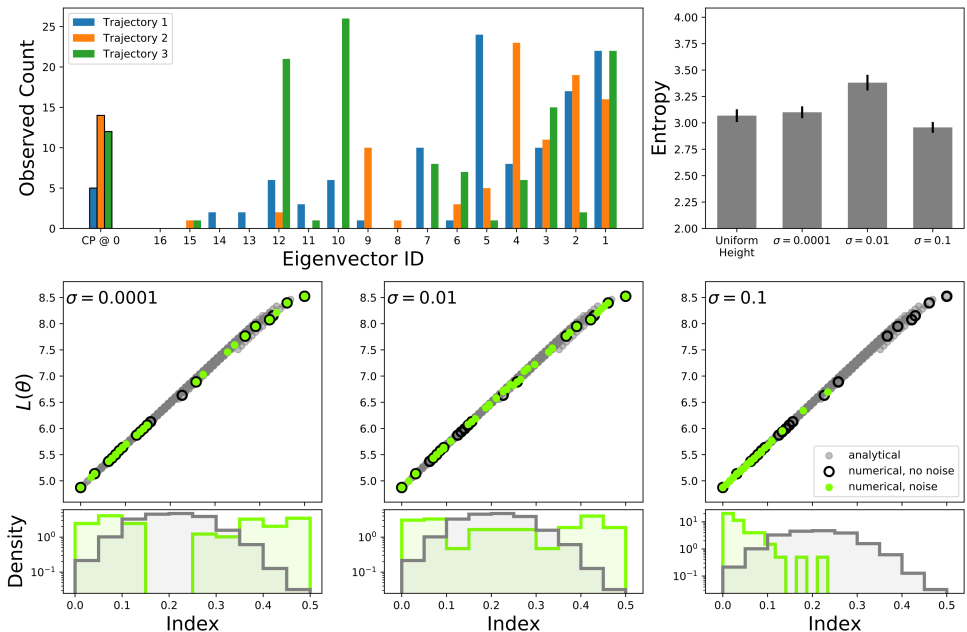}
\caption{\textbf{Additive noise with the correct magnitude reduces sampling bias}.
Top left panel: number of times an $\varepsilon$-CP that performed projection onto a given eigenvector was found.
Colors differentiate runs seeded from different optimization trajectories.
Eigenvectors are numbered in order of increasing eigenvalue.
Bars outlined in black indicate number times the critical point at zero was found.
Top right panel: entropy of distribution of eigenvector IDs (as in top left panel) for $\varepsilon$-CPs found by adding Gaussian noise with variance $\sigma^2$ to points sampled from optimization trajectories prior to executing a critical point-finding algorithm (here, Newton-MR).
Error bars indicate standard deviation, computed from bootstrapped samples ($N=100$).
Bottom row: loss and index values recovered when adding noise as in top right panel, along with log-scaled histograms of index values.
The values found by executing the same algorithm from the same trajectory without noise are indicated with empty black circles.
}
\label{fig:noise-sampling}
\end{figure*}

As discussed above, neither sampling uniformly from the trajectory nor from height on the loss surface guarantee an unbiased sample of the set $\Theta_\varepsilon$.
Indeed, the sample of critical points found using a single optimization trajectory as a seed using a Newton method is heavily biased (Fig.~\ref{fig:noise-sampling}, top left panel).
First, samples from all trajectories evinced a bias towards critical points that perform projection onto eigenvectors with large eigenvalue.
Second, individual trajectories are similarly biased towards a few other, seemingly random eigenvectors (e.g.~10 and 12 for the trajectory in green; 9 for the trajectory in orange).
We quantified this bias by taking the distribution of targeted eigenvectors (plotted as histograms in Fig.~\ref{fig:noise-sampling}, top left panel) and computing the entropy (Fig.~\ref{fig:noise-sampling}, top right panel).

To reduce this sampling bias,~\cite{dauphin2014} proposed adding noise to the values sampled from the trajectory.
We investigated whether this approach worked with Gaussian noise.
All results presented in this section were obtained using Newton-MR.
Results for Newton-TR were qualitatively similar.
We found that bias was partially reduced for appropriate choices of noise variance $\sigma^2$ (Fig.~\ref{fig:noise-sampling}, top right panel and bottom row).
For noise with low variance (Fig.~\ref{fig:noise-sampling}, bottom-left-most panel), there was no substantial change in the either the entropy or the identity of the recovered critical points.
For noise with high variance
(Fig.~\ref{fig:noise-sampling}, bottom-right-most panel), adding noise did not increase the entropy and amplified the bias towards critical points projecting onto eigenvectors with large eigenvalues.
This result is somewhat curious, since the loss values of the noise-corrupted points are higher, not lower, than the originals.
It indicates that Newton methods do not always converge to points nearby in loss value.
For intermediate values of noise variance (Fig.~\ref{fig:noise-sampling}, bottom-center panel), adding noise reduced the bias, as quantified by the entropy (bootstrap ($N=100$) means 3.05 bits, no noise, and 3.34 bits, $\sigma=0.01$; compared with Student's $t$: $t=-31.2$, $p\ll 1e-4$), and resulted in a closer match between the distributions of the indices of the true and computed CPs (see histograms in Fig.~\ref{fig:noise-sampling}, bottom row).
This value of $\sigma^2$ corresponded to a signal-to-noise ratio of 2.8 dB, compared to SNRs of 6.8 dB and 0.8 dB in the other cases.
This suggests that, to optimally reduce bias, added noise must be of sufficient magnitude to perturb the parameters but not of lower magnitude than the parameters themselves.

\subsection{Sampling Bias is Worse when Initializing Uniformly Across Iterations}

\begin{figure*}[ht]
\centering
\includegraphics[width=0.9\linewidth]{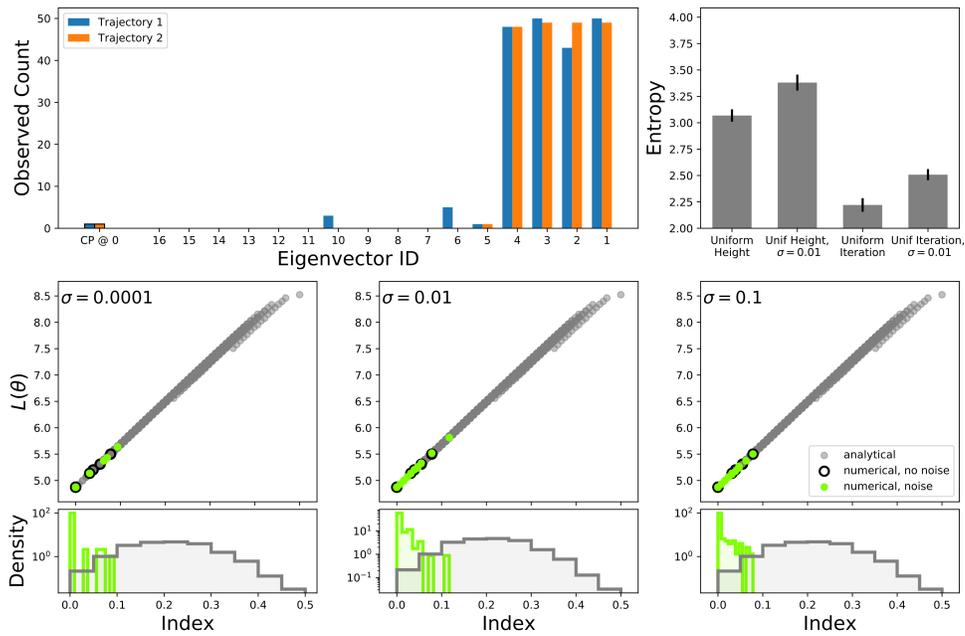}
\caption{\textbf{Uniformly sampling across iterates increases the sampling bias of critical point-finding methods}.
Top left panel: as in Fig.~\ref{fig:noise-sampling}, but for critical point-finding algorithms initialized from uniformly chosen iterates of the seed optimization trajectory.
Top right panel: as in Fig.~\ref{fig:noise-sampling}, comparing entropy across noiseless and noise-added versions of both forms of initial point sampling.
Bottom row: as in Fig.~\ref{fig:noise-sampling}.
}
\label{fig:uniform-trajectory}
\end{figure*}

Previous work sampled initial points for critical point-finding algorithms from optimization trajectories either uniformly across iterations~\cite{dauphin2014} or uniformly in height on the loss surface~\cite{pennington2017}. Note that all results reported above used the latter method.
We compared these two methods and found that sampling uniformly across iterations resulted in a heavy bias towards critical points projecting onto dominant eigenvectors (Fig.~\ref{fig:uniform-trajectory}, top left and bottom left panels).
We quantified this with the entropy of the distribution of eigenvectors onto which critical points projected, as above, and found that the entropy was significantly lower when sampling uniformly across iterations (Fig.~\ref{fig:uniform-trajectory}, top right panel).
Adding Gaussian noise to the sampled values had no effect for small values and did not fully counteract the reduction in entropy for intermediate values (Fig.~\ref{fig:uniform-trajectory}, bottom row; uniform height vs uniform trajectory: means $3.05$ bits, $2.22$ bits, $t=94.9$, $p\ll 1e-4$; uniform height vs uniform trajectory, $\sigma=0.01$: means $3.05$ bits, $2.50$ bits, $t=68.8$, $p\ll 1e-4$).

\section{Discussion}
We examined the performance of three methods to numerically find the critical points of neural network loss surfaces on a test problem for which the ground truth critical points are known.
In the absence of this ground truth and in the presence of numerical and convergence concerns, the fidelity of reported critical points to true critical points is unknown.
We found that, while all three methods are capable of finding the critical points of a deep linear autoencoder, the Newton method-based algorithms outperform direct gradient norm minimization (GNM).
As predicted by theory, GNM frequently (on 62.7\% of runs) gets stuck in local minima and requires two orders of magnitude more iterations to converge, likely due to extremely poor conditioning.

We identified a recently-proposed Newton method, Newton-MR, as a promising candidate algorithm and found that it produced solutions with fewer iterations and in less wall time than the trust region Newton method used in~\cite{dauphin2014}.
The possible applicability of quasi-Newton methods~\cite{liu1989} for large problems remains unexplored and is left to future investigation.
Newton-based homotopy methods~\cite{mehta2015} are another promising future direction, based on preliminary results in~\cite{coetzee1997}.

Numerically-recovered critical points rarely have gradient norm exactly zero, and the appropriate cutoff value for faithfully representing the loss and index values of the true critical points of a neural network is unknown.
Our results suggest that for precise recovery of loss and index values, this cutoff in the squared norm should be stricter (1e-10) than previously reported (1e-06 in~\cite{pennington2017}, unreported in~\cite{dauphin2014}), presuming that the relevant Lipschitz constants for nonlinear networks are at least as large as those for linear networks.

Given the infeasibility of calculating all of the critical points of the loss surface, the goal of critical point-finding methods should be to produce an unbiased picture of the true critical points.
We find that previously reported sampling methods based on optimization trajectories are biased towards critical points at low values of the loss and towards a random other subset of critical points, possibly determined by which critical points were closest to the optimization trajectory.
Adding noise was insufficient to remove this bias, but did mitigate it slightly.
Interestingly, we found that adding larger amounts of noise actually increased the bias towards critical points at low values of the loss.
Further work is needed to explain this phenomenon.

Taken together, our findings suggest that numerically recovering the critical points of nonlinear neural networks with high accuracy is possible, even if the problem of obtaining an even sample is still unsolved.
For questions regarding just the parts of the loss surface that a typical training trajectory passes through, relevant for answering questions about optimization, the latter is not such a serious problem.
However, it remains problematic for questions about the mathematical properties of the ensemble of critical points.
Answering these questions for a wide variety of neural networks can provide insight into which properties of the loss surface are most relevant for neural network training and generalization and by which mechanisms they arise.

\section*{Acknowledgements}

The authors would like to thank Jesse Livezey, Shariq Mobin, Jascha Sohl-Dickstein, Max Simchowitz, Levent Sagun, Yasaman Bahri, and Jeffrey Pennington for useful discussions.
Author CF was supported by the National Science Foundation Graduate Research Fellowship Program under Grant No. DGE 1752814.
Authors CF \& KB were funded by a DOE/LBNL LDRD, ‘Deep Learning for Science’, (PI, Prabhat).
NW was supported by the Google PhD Fellowship.
MRD was supported in part by the U. S. Army Research Laboratory and the U. S. Army Research Office under contract W911NF-13-1-0390.

\clearpage

\bibliography{bibliography}

\begin{thebibliography}{48}
\providecommand{\natexlab}[1]{#1}
\providecommand{\url}[1]{\texttt{#1}}
\expandafter\ifx\csname urlstyle\endcsname\relax
  \providecommand{\doi}[1]{doi: #1}\else
  \providecommand{\doi}{doi: \begingroup \urlstyle{rm}\Url}\fi

\bibitem[Advani \& Saxe(2017)Advani and Saxe]{advani2017}
Advani, M.~S. and Saxe, A.~M.
\newblock High-dimensional dynamics of generalization error in neural networks.
\newblock \emph{arXiv preprint arXiv:1710.03667}, 2017.

\bibitem[Angelani et~al.(2000)Angelani, Leonardo, Ruocco, Scala, and
  Sciortino]{angelani2000}
Angelani, L., Leonardo, R.~D., Ruocco, G., Scala, A., and Sciortino, F.
\newblock Saddles in the energy landscape probed by supercooled liquids.
\newblock \emph{Physical Review Letters}, 85\penalty0 (25):\penalty0
  5356--5359, 2000.

\bibitem[Auer et~al.(1996)Auer, Herbster, and Warmuth]{auer1996}
Auer, P., Herbster, M., and Warmuth, M.~K.
\newblock Exponentially many local minima for single neurons.
\newblock In Touretzky, D.~S., Mozer, M.~C., and Hasselmo, M.~E. (eds.),
  \emph{Advances in Neural Information Processing Systems 8}, pp.\  316--322.
  MIT Press, 1996.

\bibitem[Baldi \& Hornik(1989)Baldi and Hornik]{baldi1989}
Baldi, P. and Hornik, K.
\newblock Neural networks and principal component analysis: Learning from
  examples without local minima.
\newblock \emph{Neural Networks}, 2\penalty0 (1):\penalty0 53 -- 58, 1989.

\bibitem[Ballard et~al.(2017)Ballard, Das, Martiniani, Mehta, Sagun, Stevenson,
  and Wales]{ballard2017}
Ballard, A.~J., Das, R., Martiniani, S., Mehta, D., Sagun, L., Stevenson,
  J.~D., and Wales, D.~J.
\newblock Energy landscapes for machine learning.
\newblock \emph{Phys. Chem. Chem. Phys.}, 19:\penalty0 12585--12603, 2017.

\bibitem[Bates et~al.(2013)Bates, Haunstein, Sommese, and Wampler]{bates2013}
Bates, D.~J., Haunstein, J.~D., Sommese, A.~J., and Wampler, C.~W.
\newblock \emph{Numerically Solving Polynomial Systems with Bertini (Software,
  Environments and Tools)}.
\newblock Society for Industrial and Applied Mathematics, 2013.
\newblock ISBN 1611972698.

\bibitem[Boyd \& Vandenberghe(2004)Boyd and Vandenberghe]{boyd2004}
Boyd, S. and Vandenberghe, L.
\newblock \emph{Convex Optimization}.
\newblock Cambridge University Press, New York, NY, USA, 2004.
\newblock ISBN 0521833787.

\bibitem[Broderix et~al.(2000)Broderix, Bhattacharya, Cavagna, Zippelius, and
  Giardina]{broderix2000}
Broderix, K., Bhattacharya, K.~K., Cavagna, A., Zippelius, A., and Giardina, I.
\newblock Energy landscape of a {L}ennard-{J}ones liquid: {s}tatistics of
  stationary points.
\newblock \emph{Physical Review Letters}, 85\penalty0 (25):\penalty0
  5360--5363, 2000.

\bibitem[Broyden(1969)]{broyden1969}
Broyden, C.~G.
\newblock A new method of solving nonlinear simultaneous equations.
\newblock \emph{The Computer Journal}, 12\penalty0 (1):\penalty0 94--99, 1969.

\bibitem[Choi et~al.(2011)Choi, Paige, and Saunders]{choi2011}
Choi, S.-C.~T., Paige, C.~C., and Saunders, M.~A.
\newblock {MINRES}-{QLP}: A {K}rylov subspace method for indefinite or singular
  symmetric systems.
\newblock \emph{{SIAM} Journal on Scientific Computing}, 33\penalty0
  (4):\penalty0 1810--1836, 2011.

\bibitem[Choromanska et~al.(2014)Choromanska, Henaff, Mathieu, Arous, and
  LeCun]{choromanska2014}
Choromanska, A., Henaff, M., Mathieu, M., Arous, G.~B., and LeCun, Y.
\newblock The loss surface of multilayer networks.
\newblock \emph{CoRR}, abs/1412.0233, 2014.

\bibitem[Coetzee \& Stonick(1997)Coetzee and Stonick]{coetzee1997}
Coetzee, F. and Stonick, V.~L.
\newblock 488 solutions to the {XOR} problem.
\newblock In Mozer, M.~C., Jordan, M.~I., and Petsche, T. (eds.),
  \emph{Advances in Neural Information Processing Systems 9}, pp.\  410--416.
  MIT Press, 1997.

\bibitem[Dauphin et~al.(2014)Dauphin, Pascanu, G{\"{u}}l{\c{c}}ehre, Cho,
  Ganguli, and Bengio]{dauphin2014}
Dauphin, Y., Pascanu, R., G{\"{u}}l{\c{c}}ehre, {\c{C}}., Cho, K., Ganguli, S.,
  and Bengio, Y.
\newblock Identifying and attacking the saddle point problem in
  high-dimensional non-convex optimization.
\newblock \emph{CoRR}, abs/1406.2572, 2014.

\bibitem[Davidenko(1953)]{davidenko1953}
Davidenko, D.~F.
\newblock On a new method of numerical solution of systems of nonlinear
  equations.
\newblock \emph{Dokl. Akad. Nauk SSSR}, 88:\penalty0 601--602, 1953.

\bibitem[Dinh et~al.(2017)Dinh, Pascanu, Bengio, and Bengio]{dinh2017}
Dinh, L., Pascanu, R., Bengio, S., and Bengio, Y.
\newblock Sharp minima can generalize for deep nets.
\newblock \emph{CoRR}, abs/1703.04933, 2017.

\bibitem[Doye \& Wales(2002)Doye and Wales]{doye2002}
Doye, J. P.~K. and Wales, D.~J.
\newblock Saddle points and dynamics of {L}ennard-{J}ones clusters, solids, and
  supercooled liquids.
\newblock \emph{The Journal of Chemical Physics}, 116\penalty0 (9):\penalty0
  3777--3788, 2002.

\bibitem[Freeman \& Bruna(2016)Freeman and Bruna]{freeman2016}
Freeman, C.~D. and Bruna, J.
\newblock Topology and geometry of half-rectified network optimization.
\newblock \emph{arXiv preprint arXiv:1611.01540}, 2016.

\bibitem[Gouk et~al.(2018)Gouk, Frank, Pfahringer, and Cree]{gouk2018}
Gouk, H., Frank, E., Pfahringer, B., and Cree, M.
\newblock Regularisation of neural networks by enforcing {L}ipschitz
  continuity.
\newblock \emph{arXiv preprint arXiv:1804.04368}, 2018.

\bibitem[Griewank \& Osborne(1983)Griewank and Osborne]{griewank1983}
Griewank, A. and Osborne, M.~R.
\newblock Analysis of {N}ewton's method at irregular singularities.
\newblock \emph{SIAM Journal on Numerical Analysis}, 20\penalty0 (4):\penalty0
  747--773, 1983.
\newblock ISSN 00361429.

\bibitem[Hochreiter \& Schmidhuber(1997)Hochreiter and
  Schmidhuber]{hochreiter1997}
Hochreiter, S. and Schmidhuber, J.
\newblock Flat minima.
\newblock \emph{Neural Computation}, 9\penalty0 (1):\penalty0 1--42, 1997.

\bibitem[Izmailov \& Solodov(2014)Izmailov and Solodov]{izmailov2014}
Izmailov, A.~F. and Solodov, M.~V.
\newblock \emph{Newton-Type Methods for Optimization and Variational Problems}.
\newblock Springer International Publishing, 2014.
\newblock \doi{10.1007/978-3-319-04247-3}.

\bibitem[Jin et~al.(2017)Jin, Ge, Netrapalli, Kakade, and Jordan]{jin2018a}
Jin, C., Ge, R., Netrapalli, P., Kakade, S.~M., and Jordan, M.~I.
\newblock How to escape saddle points efficiently.
\newblock \emph{CoRR}, abs/1703.00887, 2017.

\bibitem[Jin et~al.(2018)Jin, Liu, Ge, and Jordan]{jin2018c}
Jin, C., Liu, L.~T., Ge, R., and Jordan, M.~I.
\newblock Minimizing nonconvex population risk from rough empirical risk.
\newblock \emph{CoRR}, abs/1803.09357, 2018.

\bibitem[Kehoe(1987)]{kehoe1987}
Kehoe, T.~J.
\newblock Computation and multiplicity of economic equilibria.
\newblock \emph{The Economy as an Evolving Complex System}, pp.\  147--167,
  1987.

\bibitem[Laurent \& von Brecht(2017)Laurent and von Brecht]{laurent2018}
Laurent, T. and von Brecht, J.
\newblock Deep linear neural networks with arbitrary loss: {a}ll local minima
  are global.
\newblock \emph{CoRR}, abs/1712.01473, 2017.

\bibitem[LeCun et~al.(2015)LeCun, Bengio, and Hinton]{lecun2015}
LeCun, Y., Bengio, Y., and Hinton, G.
\newblock Deep learning.
\newblock \emph{Nature}, 521:\penalty0 436 -- 444, 2015.

\bibitem[Lee et~al.(2016)Lee, Simchowitz, Jordan, and Recht]{lee2016}
Lee, J.~D., Simchowitz, M., Jordan, M.~I., and Recht, B.
\newblock Gradient descent only converges to minimizers.
\newblock In Feldman, V., Rakhlin, A., and Shamir, O. (eds.), \emph{29th Annual
  Conference on Learning Theory}, volume~49 of \emph{Proceedings of Machine
  Learning Research}, pp.\  1246--1257, Columbia University, New York, New
  York, USA, 2016. PMLR.

\bibitem[Levenberg(1944)]{levenberg1944}
Levenberg, K.
\newblock A method for the solution of certain non-linear problems in least
  squares.
\newblock \emph{Quarterly of Applied Mathematics}, 2\penalty0 (2):\penalty0
  164--168, 1944.

\bibitem[Liang et~al.(2018)Liang, Sun, Li, and Srikant]{liang2018}
Liang, S., Sun, R., Li, Y., and Srikant, R.
\newblock Understanding the loss surface of neural networks for binary
  classification.
\newblock \emph{CoRR}, abs/1803.00909, 2018.

\bibitem[Liu \& Nocedal(1989)Liu and Nocedal]{liu1989}
Liu, D.~C. and Nocedal, J.
\newblock On the limited memory {BFGS} method for large scale optimization.
\newblock \emph{Mathematical Programming}, 45\penalty0 (1-3):\penalty0
  503--528, 1989.

\bibitem[Maclaurin(2016)]{maclaurin2016}
Maclaurin, D.
\newblock \emph{Modeling, Inference and Optimization with Composable
  Differentiable Procedures}.
\newblock PhD thesis, Harvard University, April 2016.

\bibitem[McIver \& Komornicki(1972)McIver and Komornicki]{mciver1972}
McIver, J.~W. and Komornicki, A.
\newblock Structure of transition states in organic reactions. general theory
  and an application to the cyclobutene-butadiene isomerization using a
  semiempirical molecular orbital method.
\newblock \emph{Journal of the American Chemical Society}, 94\penalty0
  (8):\penalty0 2625--2633, 1972.

\bibitem[Mehta et~al.(2015)Mehta, Chen, Morgan, and Wales]{mehta2015}
Mehta, D., Chen, T., Morgan, J. W.~R., and Wales, D.~J.
\newblock Exploring the potential energy landscape of the {T}homson problem via
  {N}ewton homotopies.
\newblock \emph{The Journal of Chemical Physics}, 142\penalty0 (19):\penalty0
  194113, 2015.

\bibitem[Mehta et~al.(2018{\natexlab{a}})Mehta, Chen, Tang, and
  Hauenstein]{mehta2018a}
Mehta, D., Chen, T., Tang, T., and Hauenstein, J.~D.
\newblock The loss surface of deep linear networks viewed through the algebraic
  geometry lens, 2018{\natexlab{a}}.

\bibitem[Mehta et~al.(2018{\natexlab{b}})Mehta, Zhao, Bernal, and
  Wales]{mehta2018b}
Mehta, D., Zhao, X., Bernal, E.~A., and Wales, D.~J.
\newblock Loss surface of {XOR} artificial neural networks.
\newblock \emph{Physical Review E}, 97\penalty0 (5), 2018{\natexlab{b}}.

\bibitem[Pascanu et~al.(2014)Pascanu, Dauphin, Ganguli, and
  Bengio]{pascanu2014}
Pascanu, R., Dauphin, Y.~N., Ganguli, S., and Bengio, Y.
\newblock On the saddle point problem for non-convex optimization.
\newblock \emph{CoRR}, abs/1405.4604, 2014.

\bibitem[Pearlmutter(1994)]{pearlmutter1994}
Pearlmutter, B.~A.
\newblock Fast exact multiplication by the {H}essian.
\newblock \emph{Neural Computation}, 6:\penalty0 147--160, 1994.

\bibitem[Pennington \& Bahri(2017)Pennington and Bahri]{pennington2017}
Pennington, J. and Bahri, Y.
\newblock Geometry of neural network loss surfaces via random matrix theory.
\newblock In \emph{International Conference on Learning Representations
  (ICLR)}, 2017.

\bibitem[Roosta et~al.(2018)Roosta, Liu, Xu, and Mahoney]{roosta2018}
Roosta, F., Liu, Y., Xu, P., and Mahoney, M.~W.
\newblock Newton-{M}{R}: {N}ewton's method without smoothness or convexity.
\newblock \emph{arXiv preprint arXiv:1810.00303}, 2018.

\bibitem[Sagun et~al.(2017)Sagun, Evci, G{\"{u}}ney, Dauphin, and
  Bottou]{sagun2017}
Sagun, L., Evci, U., G{\"{u}}ney, V.~U., Dauphin, Y., and Bottou, L.
\newblock Empirical analysis of the {H}essian of over-parametrized neural
  networks.
\newblock \emph{CoRR}, abs/1706.04454, 2017.

\bibitem[Saxe et~al.(2013)Saxe, McClelland, and Ganguli]{saxe2013}
Saxe, A.~M., McClelland, J.~L., and Ganguli, S.
\newblock Exact solutions to the nonlinear dynamics of learning in deep linear
  neural networks.
\newblock \emph{CoRR}, abs/1312.6120, 2013.

\bibitem[Schmidhuber(2014)]{schmidhuber2014}
Schmidhuber, J.
\newblock Deep learning in neural networks: An overview.
\newblock \emph{CoRR}, abs/1404.7828, 2014.

\bibitem[Sommese et~al.(2005)Sommese, Verschelde, and Wampler]{sommese2005}
Sommese, A.~J., Verschelde, J., and Wampler, C.~W.
\newblock Introduction to numerical algebraic geometry.
\newblock In \emph{Solving polynomial equations}, pp.\  301--337. Springer,
  2005.

\bibitem[Trygubenko \& Wales(2004)Trygubenko and Wales]{trygubenko2004}
Trygubenko, S.~A. and Wales, D.~J.
\newblock A doubly nudged elastic band method for finding transition states.
\newblock \emph{The Journal of Chemical Physics}, 120\penalty0 (5):\penalty0
  2082--2094, 2004.

\bibitem[Wales(2004)]{wales2005}
Wales, D.
\newblock \emph{Energy Landscapes: Applications to Clusters, Biomolecules and
  Glasses}.
\newblock Cambridge University Press, 2004.

\bibitem[Wolfe(1971)]{wolfe1971}
Wolfe, P.
\newblock Convergence conditions for ascent methods. {II}: Some corrections.
\newblock \emph{{SIAM} Review}, 13\penalty0 (2):\penalty0 185--188, 1971.

\bibitem[Yao et~al.(2018)Yao, Gholami, Lei, Keutzer, and Mahoney]{yao2018}
Yao, Z., Gholami, A., Lei, Q., Keutzer, K., and Mahoney, M.~W.
\newblock Hessian-based analysis of large batch training and robustness to
  adversaries.
\newblock \emph{CoRR}, abs/1802.08241, 2018.

\bibitem[Zhou \& Liang(2017)Zhou and Liang]{zhou2017}
Zhou, Y. and Liang, Y.
\newblock Critical points of neural networks: {a}nalytical forms and landscape
  properties.
\newblock \emph{arXiv preprint arXiv:1710.11205v1}, 2017.

\end{thebibliography}
\bibliographystyle{icml2019}

\clearpage

\onecolumn

\section{Supplementary Material}
\subsection{Detailed Methods}

% All of the code used for this paper can be found at \hyperlink{https://github.com/rctn/OptimizationLandscapes/tree/autograd}.
% The figures and statistical tests can be recreated from the raw trajectory data using the Jupyter notebook ``Figure Generation.ipynb''.
% The GitHub repo can be deployed to the cloud service Binder with a single click.

\subsection{Data}

The input data to the network was a 10,000 element sample from a 16-dimensional Gaussian with mean zero and diagonal covariance matrix with entries $1 \dots 16$.
Because the analytical results~\cite{baldi1989} were derived for centered data, the data was then zero-centered to floating precision by subtracting the sample mean.

\subsection{Network and Training}

The linear autoencoder network had 16 input units, 4 hidden units, and 16 output units.
Initial parameter values were sampled from a Gaussian with variance equal to the inverse of the number of weights in each weight matrix.
The network was then trained for 10,000 epochs of full-batch gradient descent with a fixed learning rate of 0.01.
Final parameter values had losses within 5e-6 of the loss of the global minimizer (starting from losses above it by order 1).

\subsection{Back-Tracking Line Search}

The step sizes for minimizing the gradient norm objective $G$ (Equation~\ref{eq:gnm-obj-fun}) and along the Newton-MR search direction were computed using back-tracking line search.
The initial step size was 0.1 and the step size was multiplied by 0.5 when a proposed update failed to meet the update criteria (Wolfe conditions for GNM~\cite{wolfe1971}, criteria from~\cite{roosta2018} for Newton-MR).
In both cases, the free parameter for the Armijo-type condition was set to 1e-04.
In the former, the free parameter for the curvature-type condition was set to 0.9.
After a step was accepted, step size was multiplied by 2 and used as the initial step size for the next step.
Line search was terminated when multiplying the proposed step size by 0.5 resulted in no change in the value of the step size, in the number type in use.
All of our experiments used double-precision floats.

\subsection{Newton-TR}

For the Newton-TR update (Equation~\ref{eq:newton-tr}), we used 5 evenly log-spaced values of $\gamma$ from 1 to 1e-04.
We found that smaller ranges reduced convergence.
We used a step size of 0.1.

\subsection{Calculating Index and Identifying Critical Points}

To calculate index numerically, a minimum negative eigenvalue tolerance must be set.
We chose 1e-05, the square root of our criterion for the squared gradient norm.

Critical points were identified as follows.
First the un-factorized linear map that the network applies was calculated by multiplying together the input and output weights.
Near a critical point, this map should be strongly diagonally dominant, thanks to the diagonal structure of the data's true covariance matrix.
We found that simply identifying which diagonal elements were above 0.9 was sufficient to determine critical point identity: doing so did not result in identifying putative critical points that performed projection onto more than 4 eigenvectors.

\end{document}